\pdfoutput=1

\documentclass[11pt]{article}

\usepackage[]{EMNLP2022}

\usepackage{times}
\usepackage{latexsym}
\usepackage{color}

\usepackage[T1]{fontenc}

\usepackage[utf8]{inputenc}

\usepackage{microtype}

\usepackage{inconsolata}

\usepackage{multirow}
\usepackage{amsmath}
\usepackage{algorithm}
\usepackage{algpseudocode}
\usepackage{booktabs}       
\usepackage{graphicx}
\usepackage{amssymb}

\usepackage{enumitem}

\newcommand{\diff}[1]{\textcolor{magenta}{[#1]}}
\newcommand{\inc}[1]{\textcolor{teal}{[+#1]}}
\usepackage[colorinlistoftodos,prependcaption,textsize=tiny]{todonotes}
\usepackage{marginnote}

\title{CPL: Counterfactual Prompt Learning for Vision and Language Models}

\author{%
  Xuehai He$^{1}$ \,\, Diji Yang$^{1}$ \,\, Weixi Feng$^{2}$ \,\, Tsu-Jui Fu$^{2}$ \,\, Arjun Akula$^{3}$\,\, Varun Jampani$^{3}$\,\, \\ \, \textbf{Pradyumna Narayana$^{3}$\,\,Sugato Basu$^{3}$\,\, William Yang Wang$^{2}$\,\, Xin Eric Wang}$^{1}$ \\ 
  $^1$UC Santa Cruz, $^2$UC Santa Barbara, $^3$Google\\
  \texttt{\{xhe89,dyang39,xwang366\}@ucsc.edu}\\
  \texttt{\{weixifeng,tsu-juifu,william\}@ucsb.edu} \\
  \texttt{\{arjunakula,varunjampani,pradyn,sugato\}@google.com}  \\}
  
\usepackage{pdfpages}

\begin{document}
\maketitle
\begin{abstract}
Prompt tuning is a new few-shot transfer learning technique that only tunes the learnable prompt for pre-trained vision and language models such as CLIP. However, existing prompt tuning methods tend to learn spurious or entangled representations, which leads to poor generalization to unseen concepts.
Towards non-spurious and efficient prompt learning from limited examples, this paper presents a novel \underline{\textbf{C}}ounterfactual \underline{\textbf{P}}rompt \underline{\textbf{L}}earning (CPL) method for vision and language models, which simultaneously employs counterfactual generation and contrastive learning in a joint optimization framework.
Particularly, CPL constructs counterfactual by identifying minimal non-spurious feature change between semantically-similar positive and negative samples that causes concept change and learns more generalizable prompt representation from both factual and counterfactual examples via contrastive learning.
Extensive experiments demonstrate that CPL can obtain superior few-shot performance on different vision and language tasks than previous prompt tuning methods on CLIP. On image classification, we achieve a 3.55\% average relative improvement on unseen classes across seven datasets; on image-text retrieval and visual question answering, we gain up to 4.09\% and 25.08\% relative improvements across three few-shot scenarios on unseen test sets respectively.
\footnote{Our code is released at \url{https://github.com/eric-ai-lab/CPL}.}
\end{abstract}

\section{Introduction}
Pre-trained vision and language foundation models~\cite{clip,align} have shown encouraging results toward open-domain visual-concept matching.
Benefiting from prompt engineering~\cite{entailment,declaration}, where free-form text prompts are designed for specific task goals, those foundation models can be easily transferred to a wide array of tasks under zero-shot and few-shot scenarios, including image classification~\cite{imagenet}, visual question answering~\cite{how_much_can_clip}, image-text retrieval~\cite{align}, etc. 
But manually constructing prompts for vision and language models such as CLIP is a tedious, time-consuming process, which usually requires prior domain knowledge and leads to suboptimal solutions.

\begin{figure}[t]
\centering
\includegraphics[width=\linewidth]{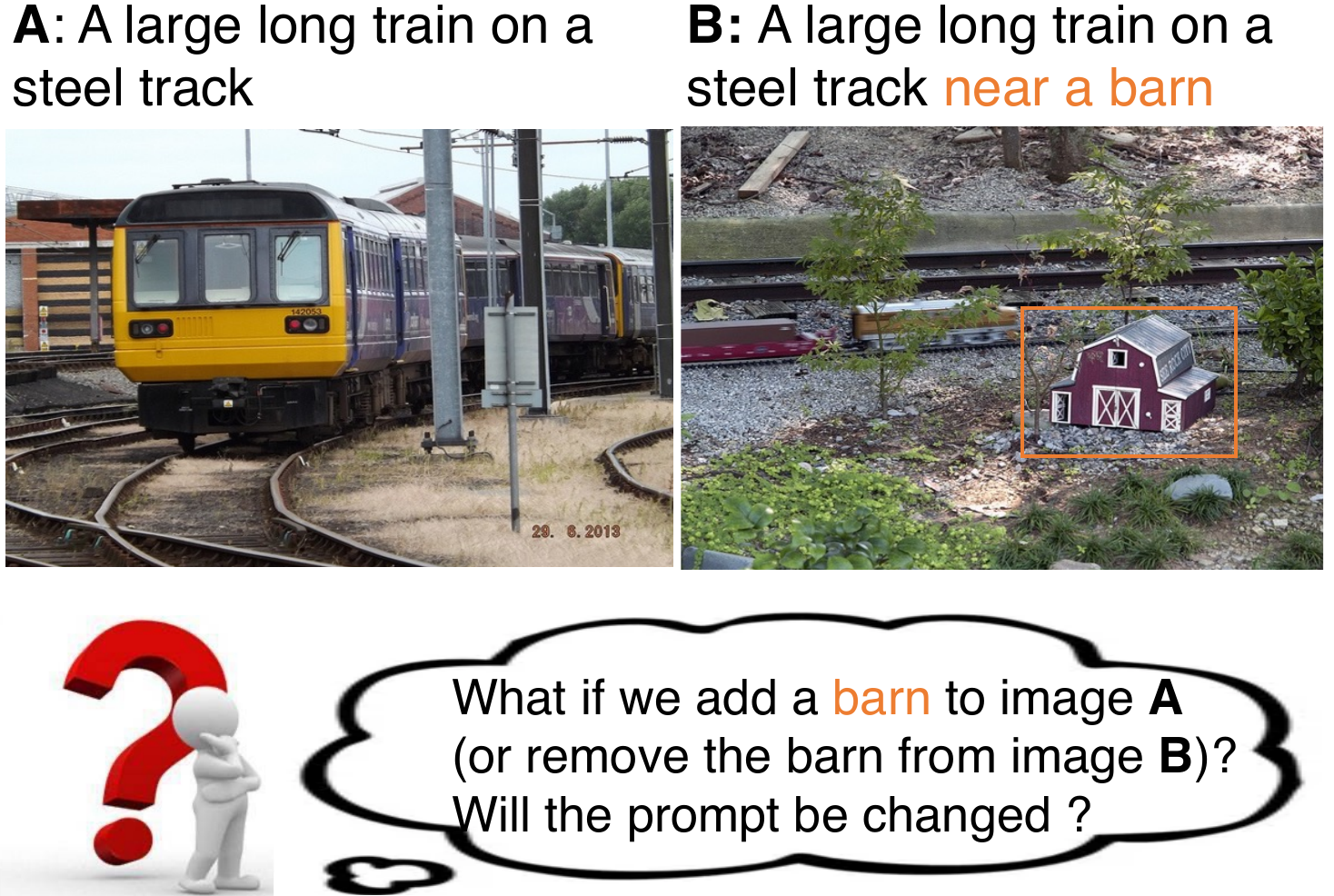}
\caption{A conceptual overview of counterfactual prompt learning. CPL constructs counterfactuals by identifying non-spurious feature change that causally causes the prompt change. In this case, the ``\textcolor{orange}{barn}'' feature is the essential cause between Prompt \textbf{A} and \textbf{B}.}
\label{fig:teaser}
\end{figure}

Prompt tuning~\cite{prompt_tuning}, on the other hand, liberates us from manual prompt engineering and automates this process. Prompt tuning methods~\cite{promptingvl,coco,cocoop} are proposed to effectively transfer CLIP to image recognition tasks after tuning a learnable prompt with a few examples of the classes.
However, those methods purely conduct empirical risk minimization (ERM) and optimize for predictive accuracy, which often produces spurious, inefficient, or entangled representations~\cite{wang2021desiderata}. 
Therefore, the generalization ability of existing prompt tuning methods for vision and language models is limited, and they often fail to transfer well to unseen classes or concepts. For example, the image classification performance of the SOTA method CoCoOp~\cite{cocoop} is similar or even degrades on unseen classes when compared with zero-shot CLIP.

Learning non-spurious representation for better generalization requires disentangling features that causally determine the prompts. One solution is counterfactual reasoning.
Counterfactual (``counter to the facts'') is a concept that describes the human capacity to learn from limited prior experiences by imagining the outcome of an alternative action that could have been taken.
So we can do counterfactual intervention by asking ``what if ...'' questions in prompt learning. For example, as shown in Figure~\ref{fig:teaser}, a change in the visual feature of the barn would cause the label to change (if we view the two prompts as two labels).

Therefore, we introduce a new causality-based approach, \underline{\textbf{C}}ounterfactual \underline{\textbf{P}}rompt \underline{\textbf{L}}earning (CPL), for non-spurious and efficient prompt learning.
First, we introduce a text-based negative sampling strategy to discover the most semantically-similar negative sample based on text similarity.
Then we generate a counterfactual example by identifying minimal non-spurious feature change between semantically-similar positive and negative samples that causally causes prompt change. Finally, we adopt contrastive learning in the joint optimization framework (with counterfactual construction) to tune the learnable prompts using both factual and counterfactual examples.
The causally fine-tuned prompts will eventually guide vision-and-language foundation models to distinguish images from unseen concepts, thereby improving the generalization ability of prompt learning.

We extensively evaluate CPL using seven standard datasets for image classification, two for image-text-retrieval, and one for visual question answering (VQA). We show that CPL outperforms the baseline on all three tasks: on image classification, our method achieves $3.55\%$ average relative improvement on unseen classes across the seven datasets in terms of accuracy; on image-text retrieval, our method improves the most ($4.09\%$ relative improvement in terms of Recall@1) when using $0.5\%$ of total training instances on MSCOCO~\cite{coco} and Flickr30K~\cite{flickr}; on VQA, we gain up to $25.08\%$ relative improvement on the VQAv2~\cite{vqav2} dataset.

Our main contributions are summarized below:
\begin{itemize}
\item We introduce \underline{\textbf{C}}ounterfactual \underline{\textbf{P}}rompt \underline{\textbf{L}}earning (CPL), a task-agnostic causality-based prompt learning method to effectively transfer CLIP to unseen concepts for different downstream tasks.
\item We propose a text-based negative sampling strategy, where we compute BERTScore~\cite{zhang2019bertscore} between text prompts, based on which we sample the most semantically-similar negative images.
\item We introduce a optimization framework that simultaneously constructs counterfactuals by identifying minimal non-spurious feature change, and learns the generalized prompt representation from both factual and counterfactual examples. 
\item We conduct extensive experiments on image classification, image-text retrieval, and visual question answering, and validate the superiority of CPL to existing prompt tuning methods in transferring effectiveness on unseen concepts.
\end{itemize}

\section{Related Work}
\paragraph{Vision-and-Language Models.~}
Vision-and-Language models pre-trained on large-scale image-text pairs have demonstrated great potential in multimodal representation learning~\cite{align,flip,florence}. Among them, the representative CLIP~\cite{clip} benefits from 400M curated data and defines various prompt templates to carry out zero-shot image classification. However, those prompts still require hand-crafted designs. In this work, we automatically learn task-agnostic and task-relevant prompts without human priors. In addition, by considering the counterfactual examples, we can further improve various vision-and-language tasks, including visual question answering and image-text retrieval in a few-shot scenario.

\paragraph{Prompt Tuning.~}
Many works focus on learning from discrete natural language prompts, e.g., AutoPrompt~\cite{shin2020autoprompt} elicits knowledge from language models with automatically generated discrete prompts. Lately, many other works~\cite{coop, cocoop} directly tune prompts in continuous vector forms. ~\citet{prompt_q_learning} introduces Q-Learning to optimize the soft prompt.  P-Tuning v2~\cite{ptuningv2} shows that continuous prompt tuning achieves the same performance as fine-tuning in various settings. Prompt tuning also receives great interest in the computer vision domain. For example, CoOp proposes a continuous prompt optimization strategy to avoid prompt design.  CoCoOp~\cite{cocoop} extends CoOp by further learning an instance-conditional network to generate an input-conditional token for each image. However, these methods trained with empirical risk minimization (ERM) may learn to rely on correlations between class labels and spurious attributes by minimizing average training error~\cite{correct_contrast}. They usually learn spurious, inefficient, and entangled representation, lacking generalization ability to unseen scenarios. 

\paragraph{Counterfactual Reasoning.~}
A number of recent works have investigated  generating counterfactual images \cite{besserve2020counterfactuals}, or counterfactual text in specific language domains (e.g., court view~\cite{wu2020biased}, dialogue generation~\cite{zhu2020counterfactual}, Natural Language Inference~\cite{kaushik2019learning, semantically_robust_optimization}, named entity recognition~\cite{zeng2020counterfactual}); On the vision end, ~\citet{causal_pose_estimator} proposes to add intervention over the changed domain on images during the data-generation process and steer the generative model to produce counterfactual features to augment the training process. ~\citet{causalvqa} uses automated semantic image manipulations to generate synthetic data to make models more robust against spurious correlations; On the vision and language end, ~\citet{counterfactual_vqa} proposes to generate counterfactual VQA samples by masking critical objects in images or words in questions to augment the training data and gain a huge improvement on the VQAv2 dataset.  ~\citet{mutant} proposes  template-based counterfactual image augmentation methods. ~\citet{counterfactual_vln} proposes a novel training strategy for visual language navigation that dynamically generates counterfactuals to account for unseen scenarios.
To our best knowledge, CPL is the first to apply counterfactual generation to prompt-based few-shot learning for vision and language models.

\paragraph{Few-shot Learning.}
Recently, many few-shot and efficient learning methods on vision~\cite{PEViT} and language~\cite{efficient_language_learning} tasks have been widely studied. At the same time, like CLIP, several different few-shot learners were proposed. GPT~\cite{gpt3}, as a strong few-shot learner, is capable of performing a new language task by learning from only a few training instances.  Frozen~\cite{frozen} is developed based on GPT and made into a multimodal few-shot learner by expanding the soft prompting to include a collection of images and text. Their 
method demonstrates strong few-shot capabilities on visual question answering and image classification tasks. Similarly, CoCa~\cite{coca} is pre-trained from scratch and end-to-end using both web-scale data and annotated
images by considering all labels as text, therefore unifying supervision for learning representations through natural language. It can achieve state-of-the-art performance with few-shot transfer or by minimal task-specific adaptation on a wide range of downstream vision-and-language tasks, including visual recognition, multimodal understanding, crossmodal retrieval, and image
captioning. SimVLM~\cite{simvlm} is pre-trained with prefix language modeling on datasets with weak supervision. It exhibits its efficacy on few-shot captioning tasks. Even though all these models mentioned above can already achieve improvement on some few-shot tasks, how to exploit their few-shot reasoning ability using limited training examples still deserves the effort. In this work, we study this direction via the lens of prompt learning utilizing CLIP as a starting point.

\begin{figure*}[t]
\centering
\includegraphics[width=\linewidth]{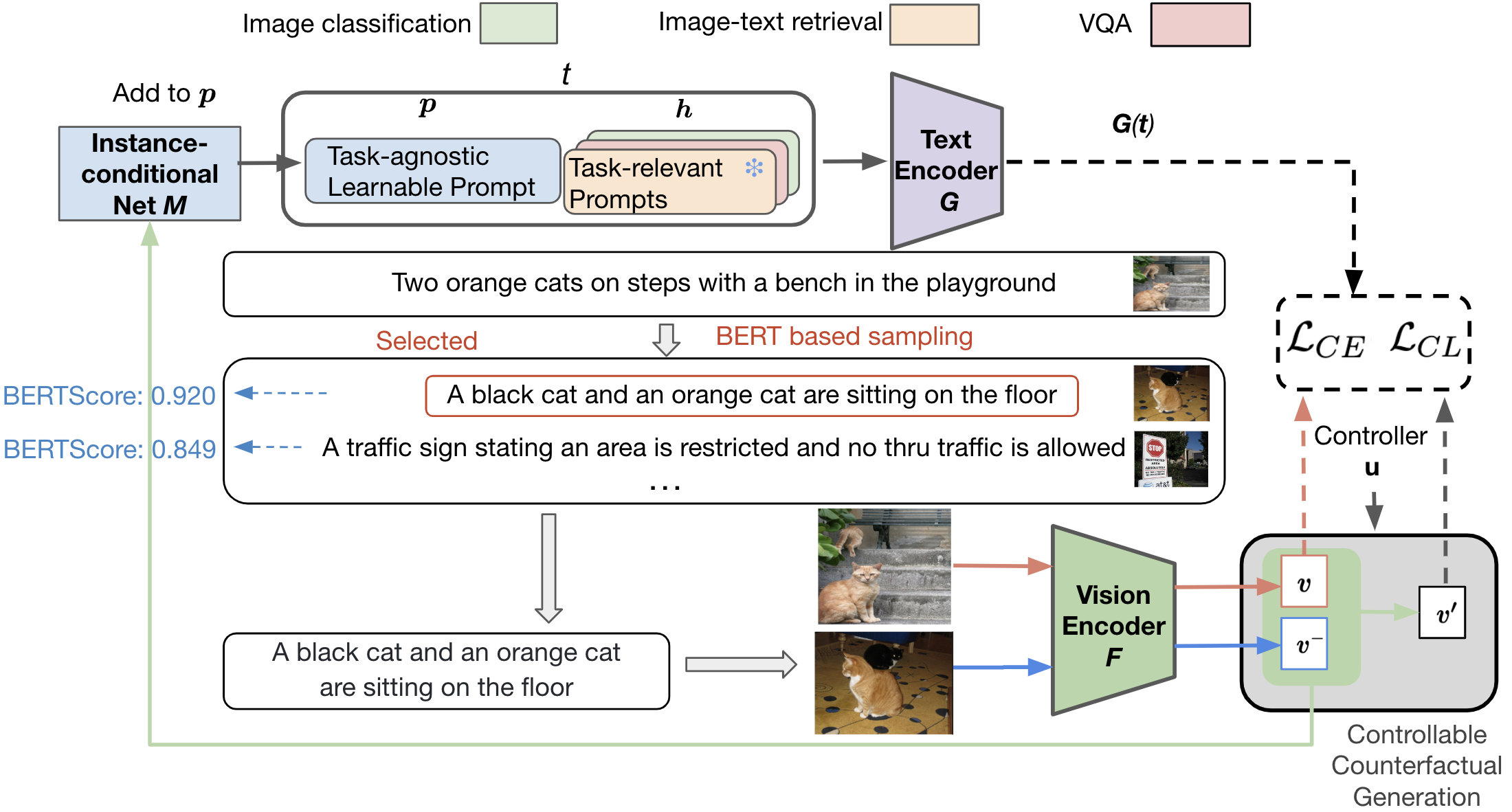}
\caption{The counterfactual prompt learning framework. We freeze the vision encoder $F$ and the text encoder $G$, and only optimize the task-agnostic prompts and the instance-conditioned net $M$ (blue blocks). Please refer to Section~\ref{sec:overview} for the explanation.
}
\label{fig:overview}
\end{figure*}

\section{Counterfactual Prompt Learning}
\label{sec:method}
\subsection{Problem Formulation}
Our goal is to learn generalizable prompt representation with limited data. The prompt in CLIP is divided into two parts: task-agnostic prompt $\boldsymbol{p}$
and task-relevant prompt $\boldsymbol{h}$.
Task-agnostic prompt $\boldsymbol{p}$ is learned end-to-end automatically. The set of task-relevant prompts $\mathbb{H}=\left\{\boldsymbol{h}_{0}, \boldsymbol{h}_{1}, \ldots, \boldsymbol{h}_{C}\right\}$ is mapped from the label space $\mathbb{Y}$ with some predefined rules hinging on the task type, where $C$ is the total number of classes. The final prompt $\boldsymbol{t}_c$ is the concatenation of the task-agnostic prompt and the task-relevant prompt fed into CLIP's text encoder:
$\boldsymbol{t}_{c}=[\boldsymbol{p}, \boldsymbol{h}_{c}]$.

Existing works to this problem~\cite{coop,cocoop} propose to first extract visual feature $\boldsymbol{v}$ of each input image by feeding it into CLIP’s vision encoder $F$; and text embeddings are generated by feeding $\left\{\boldsymbol{t}_{c}\right\}_{c=1}^{C}$ into the CLIP’s text encoder  $G$.
The probability of $i$-th class is computed as
\begin{equation}
p(\boldsymbol{t}_{i}\mid \boldsymbol{x})=\frac{e^ \frac{<G\left(\boldsymbol{t}_{i}\right), \boldsymbol{v}>}{ \tau}}{\sum_{c=1}^{C} e^ \frac{<G\left(\boldsymbol{t}_{c}\right), \boldsymbol{v}>}{ \tau}},
\label{eq:1}
\end{equation}
where $\tau$ is the temperature parameter, $<\cdot>$ denotes the cosine similarity. Cross-entropy loss is then minimized and the gradients can be back-propagated via the text encoder $G$ to update the learnable prompt representation $\boldsymbol{p}$. During training, the weights of CLIP always remain frozen. During inference, Eq.~\ref{eq:1} is used to compute the probability for each class.

\subsection{Method Overview}\label{sec:overview}
An  overview of the Counterfactual Prompt Learning (CPL) framework is shown in Figure~\ref{fig:overview}. For pre-processing, we construct task-relevant prompts for all training samples. 
The goal is to optimize the task-agnostic prompt $\boldsymbol{p}$.\footnote{Together with the instance-conditional net $\boldsymbol{M}$ as introduced in \citet{cocoop}. For simplicity, we will only use $\boldsymbol{p}$ hereafter as $\boldsymbol{p}$ and $\boldsymbol{M}$ are always optimized together.}
During training, given a positive image-prompt pair, we first perform \emph{text-based negative sampling} to find the most semantically-similar negative sample based on text similarity scores. 
Then we adopt a \emph{controllable counterfactual generation} strategy to construct the counterfactual from the positive and negative samples in the visual feature space. 
Finally, we perform contrastive learning using both generated counterfactual image features and factual image features in a joint optimization framework to fine-tune the task-agnostic prompt $\boldsymbol{p}$, allowing the model to understand non-spurious semantic information and learn generalized prompt representations.

\subsection{Controllable Counterfactual Generation}\label{sec:generation}
\begin{figure}[t]
\centering
\includegraphics[width=\linewidth]{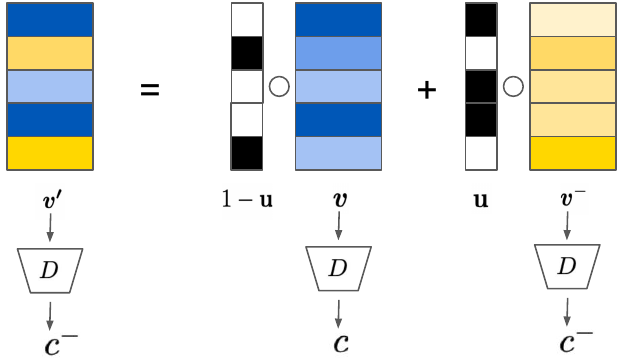}
\caption{Counterfactual generation process. $\boldsymbol{v}$ and $c$ are the positive image feature and label, while $\boldsymbol{v}^-$ and $c^-$ are the negative image feature and label. $\circ$ is element-wise multiplication. By mixing $\boldsymbol{v}$ and $\boldsymbol{v}^-$, the counterfactual image feature $\boldsymbol{v'}$ is predicted as a negative label $c^-$ by the discriminator $D$. $\mathbf{u}$ is minimized so a minimal change to the positive image feature $\mathbf{u}$ is captured here to causally change the label.}
\label{fig:generation}
\end{figure}

By viewing image feature $ \boldsymbol{v}$ as a potential cause of the label, a non-spurious feature shall be a sufficient cause of the label. So we would like to generate counterfactuals by identifying minimal non-spurious feature change that causes the label change.
The illustration of the counterfactual construction process is shown in Figure~\ref{fig:generation}.
Given positive image features $\boldsymbol{v}$ and negative image features $\boldsymbol{v^-}$, we can generate negative counterfactual image features $\boldsymbol{v'}$ as below:
 \begin{equation}
 \boldsymbol{v'} =(1-\mathbf{u}) \circ  \boldsymbol{v} + \mathbf{u}\circ \boldsymbol{v}^{-},
 \label{generation}
 \end{equation}
where $\circ$ is the element-wise multiplication and $\mathbf{u}$ is the parameter controlling the amount of negative image feature that replaces the positive image feature. 
The negative image features are extracted from those images similar to the original image at the semantic level, which we will introduce in Section~\ref{sec:nagative_sampling}.

To capture the non-spuriousness, we would like to construct counterfactuals by replacing essential non-spurious features only. This can be achieved by minimizing the amount of feature change $\mathbf{u^*}$ to the original image that can causally incur label change:
\begin{equation}
\begin{array}{cl}
\underset{ \mathbf{u}^*}{\operatorname{minimize}} &\|\mathbf{u}^*\|_{1} \\
\text { s.t. } & \mathbf{u}^*=\arg \underset{\mathbf{u}}{\max} D_{c^-}(\boldsymbol{v'}).
\end{array}
\label{eq:min-u}
\end{equation} 

Given the factual and counterfactual features $\boldsymbol{v}$ and $\boldsymbol{v'}$, we aim to learn the prompt that can help CLIP better align visual features $\boldsymbol{v}$ and textual features $G(\boldsymbol{t})$ with same semantic meanings. This can be achieved by maximizing the mutual information (MI) between $\boldsymbol{v}$ and $G(\boldsymbol{t})$. Therefore, by minimizing the InfoNCE loss~\cite{infonce}, we can maximize the lower bound on MI$(\boldsymbol{v},G(\boldsymbol{t}))$.
To this end, we define the contrastive objective function based on the InfoNCE estimator following~\citet{supervised_contrastive_learning}:
\begin{equation}
\mathcal{L}_{CL}(\boldsymbol{p}, \mathbf{u}^*) = -log(\frac{e^{\frac{S(\boldsymbol{v},G(\boldsymbol{t}))}{ \tau}}}{ e^{ \frac{S(\boldsymbol{v}, G(\boldsymbol{t}))}{\tau}}+ e^{\frac{S(\boldsymbol{v'}, G(\boldsymbol{t}))}{ \tau}}}),
\label{eq:cl}
\end{equation}
where
$S\left(\cdot, \cdot\right) $
is normally the cosine similarity function and $\tau$ is the temperature value.

\subsection{Text-based Negative Sampling}
\label{sec:nagative_sampling}
We then discuss how to perform negative sampling for constructing counterfactual features. 
As suggested in~\citet{contrastive_learning_hard_sampling}, good negative samples have different labels and are difficult to be distinguished from an anchor point, while their semantic representations are close~\citep{suresh2021not}.
Since not all negative samples can serve as useful negatives~\cite{chuang2020debiased}, indiscriminate leverage of these data may harm model robustness and algorithm efficiency.
Therefore, during training, in each batch, we only utilize the most 
semantically-similar one to generate counterfactual image features. Other image samples are filtered out.

Semantic concepts may be highly complex in the visual representations, and thus it is hard to directly measure semantic similarity in the visual space. While language is more expressive and naturally preserves semantic meanings.
Therefore, we propose a text-based negative sampling method. We first measure the text similarity between prompts with BERTScore~\cite{zhang2019bertscore}, which computes pairwise cosine similarity between reference sentences and candidate sentences using BERT contextual embedding~\citep{devlin-etal-2019-bert}. We compute a similarity matrix with the value of each element being:
\begin{equation}
\operatorname{sim}({i, j}) = \operatorname{BERTScore}(\boldsymbol{h}_i, \boldsymbol{h}_j).
\label{similarity}
\end{equation}
Denote $\mathcal{B}$ as the collection of sampled instances. During training, each prompt $ \boldsymbol{h}_c\in \mathcal{B}$ ($1 \leq c \leq C$, where $C$ is the size of sampled instances) can be treated as a query. Given a query prompt $\boldsymbol{h}_q$, its most semantically similar prompt (the one with the highest BERTScore) $\boldsymbol{h}_k$ is searched from $\mathcal{B}$.
Then we use the CLIP vision encoder to obtain the features of the corresponding positive and negative images $\boldsymbol{v}$ and $\boldsymbol{v}^{-}$.

\subsection{Joint Optimization}\label{sec:optimization}
In addition to the contrastive learning loss as introduced in Eq.~\ref{eq:cl}, we also adopt the standard cross-entropy loss for training:
\begin{equation}
\mathcal{L}_{\mathrm{CE}}(\boldsymbol{p})=-\sum_{c} \boldsymbol{y}_{c} \log p\left(\boldsymbol{t}_{c} \mid \boldsymbol{x}\right),
\label{eq:ce}
\end{equation}
where $\boldsymbol{y}_c$ denotes the one-hot ground-truth annotation of the label. We treat all downstream tasks in this work as classification tasks, where the model predicts if the image and text prompt pair is matched or not. 

Then the task-agnostic prompt $\boldsymbol{p}$ is learned by minimizing the weighted combination of contrastive learning loss and cross-entropy loss:
\begin{equation}
   \mathcal{L}(\boldsymbol{p})= \mathcal{L}_{CE}(\boldsymbol{p})+\lambda \cdot \mathcal{L}_{CL}(\boldsymbol{p}, \mathbf{u}^*),
   \label{eq:loss}
\end{equation}
where $\lambda$ determines the weight of $\mathcal{L}_{CL}$.

\begin{algorithm}[t]
\begin{small}
	\caption{Counterfactual Prompt Learning}
\label{vqa_alg}
	\begin{algorithmic}[1]
	\State $\mathbb{X}$: image space
	\State $\mathbb{Y}$: label space
	\State $\boldsymbol{h}_{c}$: task-relevant prompt for the $c$-th class
	\State $\mathbb{H}$: the set of task-relevant prompts
	\State $\boldsymbol{p}$: the task-agnostic prompt 
	\State $\boldsymbol{v}$: image features
	\State $\boldsymbol{v}^-$: negative image features
	\State $\mathbf{u}$: parameter controls the generation of counterfactual image features
		\Function {$\mathcal{\textcolor{blue}{CPL}}$}{$\mathbb{X}, \mathbb{Y}$}
		\State $\mathbb{H}\leftarrow \mathbb{Y}$  
		\State $\boldsymbol{t}_{c}\leftarrow[\boldsymbol{p}, \boldsymbol{h}_{c}]$
		\For{each $i,j$}  
		\State $\operatorname{sim}({i, j}) = \operatorname{BERTScore}(\boldsymbol{h}_i, \boldsymbol{h}_j)$~\Comment{Eq.~\ref{similarity}} 
		\EndFor
		\For{$q$ in the batch}  
		\State $\boldsymbol{v} \leftarrow \boldsymbol{v_q}$
		\State Find the index $k$ that maximize $\operatorname{sim}({q, k})$ with the given index $q$
		\State $\boldsymbol{v}^- \leftarrow \boldsymbol{v_k}$
		\State Generate counterfactual image features 
		\label{step:generation}
		\Comment{Eq.~\ref{generation}}
		\State $\mathcal{L}_{CE} \leftarrow$ cross-entropy loss~\Comment{Eq.~\ref{eq:ce}}
		\State $\mathcal{L}_{CL} \leftarrow$ contrastive loss~\Comment{Eq.~\ref{eq:cl}} 
		\State Update $\boldsymbol{p}$ and $\mathbf{u}$ with the joint optimization loss~\Comment{Eq.~\ref{eq:loss}} 
		\EndFor
		\label{step:optimizing}
		\EndFunction
	\end{algorithmic}
	\end{small}
\end{algorithm}

In fact, we can seek to put Eq.~\ref{eq:min-u} and Eq.~\ref{eq:loss} in a single-stage optimization framework. 
The intuition is that we generate counterfactual image features with minimal feature change that can maximize the negative prediction probability, and at the same time, utilize contrastive learning to learn the prompt that can guide CLIP to explicitly distinguish between factual images and counterfactual images. 
Putting all pieces together, we have:
\begin{equation}
\begin{array}{cl}
\underset{\boldsymbol{p}, \mathbf{u}^*}{\operatorname{minimize}} & \mathcal{L}_{CE}(\boldsymbol{p}) +\lambda \cdot \mathcal{L}_{CL}(\boldsymbol{p}, \mathbf{u}^*) + \|\mathbf{u}^*\|_{1} \\
\text { s.t. } & \mathbf{u}^*=\arg \underset{\mathbf{u}}{\max} D_{c^-}(\boldsymbol{v'}) \\
\text{where ~}  \boldsymbol{v'} &= (1-\mathbf{u}) \circ  \boldsymbol{v} + \mathbf{u}\circ \boldsymbol{v}^{-}.
\end{array}
\label{eq:6}
 \end{equation}
In Eq.~\ref{eq:6}, the gradients can be back-propagated all the way through the text encoder $G$ to the task-agnostic prompt, making use of the rich knowledge encoded in the pre-trained CLIP model to optimize the prompt.

Algorithm~\ref{vqa_alg} presents the learning algorithm of CPL. In summary, given few input training samples $\left\{\left(x_{1}, y_{1}\right), \ldots,\left(x_{n}, y_{n}\right)\right\}$, CPL consists of three main steps:
(1) compute the similarity matrix between different text prompts within the sampled batch;
(2) generate counterfactual image features;
(3) optimize $\boldsymbol{p}$ and $\boldsymbol{u}$ with contrastive learning loss and cross-entropy loss.

\subsection{Task-relevant Prompt Construction}\label{sec:task-relevant}
We construct task-relevant prompts $\mathbb{H}$ for image classification, image-text retrieval, and visual question answering, respectively. For image classification, the prompts are class labels for each task; for image-text retrieval, captions for each image are adopted as prompts; for visual question answering, we first use a pre-trained generative T5 model~\cite{t5} to convert the question-answer pairs into declarative sentences referring to the VQA prompt generation method proposed in~\citet{vqa_prompt}. Then, motivated by~\citet{chain_of_thought}, we add additional category information into the prompt generated from templates based on the question type to help the model perform intermediate reasoning steps. Specifically, we add ``The question is asking about others'' for \emph{Other} questions before the generated declarative sentence. In a similar vein, ``The question is asking about yes or no'' and ``The question is asking about numbers'' are added for \emph{Yes/No} and \emph{Number} questions.

\begin{table*}[t]
 \resizebox{\linewidth}{!}{
  \centering
  \setlength{\tabcolsep}{3pt}
  \begin{tabular}{llllllllll}
    \toprule
      Classes  &Method & SUN397& Caltech101 & ImageNet & OxfordPets & StanfordCars & Flowers102 & Food101 & Average\\
    \midrule
    \multirow{3}{*}{Seen} 
        &  CLIP & 69.40& 96.51& 72.46& 91.33 & 74.85 & 72.17 & 90.12& 80.98\\
        & CoCoOp & 79.08 \inc{13.95}& 97.66 \inc{1.19}& 76.01 \inc{4.90}& 95.18 \inc{4.22}& 70.91 \diff{-5.26} & \textbf{94.65} \inc{31.15}& 90.67 \inc{0.61} & 86.31 \inc{6.58}\\
        & CPL (ours)& \textbf{81.05} \inc{16.79}& \textbf{97.70} \inc{1.23}& \textbf{78.81} \inc{8.76} &\textbf{96.69} \inc{5.87} & \textbf{75.51} \inc{0.88} & 93.91 \inc{30.12} & \textbf{93.01} \inc{3.21} & \textbf{88.10} \inc{8.79} \\
    \midrule
    \multirow{3}{*}{Unseen} 
        &  CLIP & 75.40& 94.10& 68.09& 97.04& 74.95& \textbf{77.87}& 91.30&82.68\\
        & CoCoOp & 76.83 \inc{1.90} & 93.92 \diff{-0.19} & 70.44 \inc{3.45} &97.78 \inc{0.76} & 73.09 \diff{-2.48} & 69.24 \diff{-11.08} & 91.53 \inc{0.25} &81.83 \diff{-1.02} \\
        & CPL (ours) & \textbf{80.19} \inc{6.35} &\textbf{94.94} \inc{0.89} & \textbf{73.17} \inc{7.46} & \textbf{98.81} \inc{1.82} & \textbf{78.90} \inc{5.27} & 72.30 \diff{-7.15} & \textbf{93.44} \inc{2.34} &\textbf{84.54} \inc{2.25} \\
    \bottomrule
  \end{tabular}}
  \caption{Result comparison between CPL and CoCoOp~\cite{cocoop} on seen and unseen classes across seven image classification datasets in terms of accuracy (\%) under the few-shot setting.
  The relative difference (\%) compared with CLIP is reported in color. 
  }
 \label{tab:classification}
\end{table*}

\begin{table}[t]
 \resizebox{\columnwidth}{!}{
  \centering
  \setlength{\tabcolsep}{3pt}
  \begin{tabular}{lllll}
    \toprule
   Training data used & Method & {Flickr30k}& {MSCOCO}& Average \\
   \midrule
   0 & CLIP & 83.00 & 53.35&68.18\\
    \hline
   \multirow{2}{*}{0.5\%} 
    & CoCoOp & 82.40 \diff{-0.72} &55.55 \inc{4.12}&68.98 \inc{1.17}\\
    & CPL (ours) & \textbf{85.64} \inc{3.18} &\textbf{57.91} \inc{8.55}&\textbf{71.78} \inc{5.28}\\
   \midrule
   \multirow{2}{*}{1\%} 
    & CoCoOp & 84.80 \inc{2.17}&56.62 \inc{6.13}&70.71 \inc{3.71}\\
    & CPL (ours) & \textbf{86.91} \inc{4.71} &\textbf{58.43} \inc{9.52}&\textbf{72.67} \inc{6.59}\\
   \midrule
   \multirow{2}{*}{3\%} 
    & CoCoOp & 85.90 \inc{3.49}&58.08 \inc{8.87}&71.99 \inc{5.59}\\
    & CPL (ours) & \textbf{87.74} \inc{5.71} &\textbf{59.96} \inc{12.39}&\textbf{73.85} \inc{8.32}\\
    \bottomrule
  \end{tabular}}
  \caption{Result comparison between CPL and CoCoOp on two image-text retrieval datasets, Flickr30k~\cite{flickr} and MSCOCO~\cite{coco}, on the unseen test sets in terms of Recall@1 (\%).  The relative difference (\%) over CLIP is reported in color.}
 \label{tab:ir}
\end{table}

\begin{table}[t]
 \resizebox{\columnwidth}{!}{
  \centering
  \setlength{\tabcolsep}{3pt}
  \begin{tabular}{lll}
    \toprule
    Training data used & Method & VQAv2 \\
    \midrule
    0 &{CLIP}&11.83\\
    \midrule
     \multirow{3}{*}{0.5\%} & {CoCoOp}& 27.98 \inc{136.52}\\
     & {CPL w/o. Category Information}& 31.68 \inc{167.79}\\
     & {CPL }& \textbf{33.39} \inc{182.25} \\
     \midrule
      \multirow{3}{*}{1\%} & {CoCoOp}& 28.51 \inc{141.00}\\
     & {CPL w/o. Category Information}& 34.70 \inc{193.32}\\
     & {CPL }& \textbf{35.66} \inc{201.44}\\
     \midrule
      \multirow{3}{*}{3\%} & {CoCoOp}& 30.18 \inc{155.11}\\
     & {CPL w/o. Category Information}& 35.41 \inc{199.32}\\
     & {CPL }& \textbf{36.32} \inc{207.02}\\
    \bottomrule
  \end{tabular}}
 \caption{Result comparison on the VQAv2 dataset~\cite{vqav2} in terms of accuracy (\%). The relative improvements over CLIP are reported in color. Incorporating category information into task-relevant prompts can further improve the performance.}
 \label{tab:vqa}
\end{table}

\section{Experiments}
\subsection{Tasks and Datasets}
\paragraph{Image Classification.~}
We employ seven publicly available image classification datasets used in CLIP: SUN397~\cite{sun397}, Caltech101~\cite{caltech}, ImageNet~\cite{imagenet}, OxfordPets~\cite{oxfordpet}, StandfordCars~\cite{standfordcars}, Flowers102~\cite{flower}, and Food101~\cite{food101}. These datasets constitute a comprehensive benchmark, which covers a diverse set of vision tasks
including the classification of generic objects, fine-grained image recognition, action classification, etc. To evaluate the generalization ability of methods, we split those datasets into seen and unseen classes. Only images in the seen classes will be used for training. The setting follows the few-shot evaluation protocol in CLIP, where we use 16 shots for training and full test sets for testing.

\paragraph{Image-Text Retrieval.~} 
We consider two datasets for image-text retrieval: MSCOCO~\cite{coco} and Flickr30K~\cite{flickr}. We adopt the
widely used Karpathy split~\cite{karpathy2015deep} for both the MSCOCO and Flickr30K datasets, where MSCOCO contains 113/5K/5K for train/validation/test.
Flickr30K contains 29K/1K/1K images for train/validation/test. We construct few-shot setting subsets for both CoCoOp and CPL by taking $0.5\%$, $1\%$, and $3\%$ of training instances. We train the model with the subsets and evaluate its performance on the complete
test set. We use Recall at 1 (R@1) as the default evaluation metric.

\paragraph{Visual Question Answering.~}
VQAv2~\cite{goyal} is an extended dataset from the VQA~\cite{vqa} dataset. The questions are categorized
into three types: \emph{Number}, \emph{Yes/No}, and \emph{Other}. We set up the experiments following~\citet{bottom}, which treats visual question answering as a classification problem: for each question, the model picks the corresponding answer from a given set of predefined most frequent candidate answers and matches it with the image. The questions are first converted into a masked template using the pre-trained T5 model and predefined rules. The infilled template along with the questions will be turned into prompts that naturally connect questions and answers. The model will predict whether the given prompt and image pairs are matched. We construct the few-shot setting by taking $0.5\%$, $1\%$, and $3\%$ instances for training.

\subsection{Implementation Details}
\paragraph{Baselines.~} 
We mainly compare CPL with CoCoOp~\cite{cocoop}, one of the earliest prompt tuning methods proposed for vision-and-language pre-trained models.
CoCoOp considers each input image and injects the learnable instance-aware tokens into the context vectors as the final prompt. For a fair comparison, both CPL and CoCoOp adopt CLIP~\cite{clip} as the pre-trained vision-and-language backbone and are compared with respect to their relative improvements over zero-shot CLIP.

\paragraph{Prompt Tuning.~} The task-agnostic prompt is randomly initialized from a zero-mean Gaussian distribution with the standard deviation $0.02$, where we set length $L=4$ by default.  For vision and language tasks, in contrast to image classification, where an image is labeled by a category, the task-relevant prompts comprise more fine-grained details, usually a sentence. We here similarly tokenize the whole sentence using the CLIP word embedding~\cite{clip}, and feed the tokenized results to the text encoder with task-agnostic prompt vectors, to generate the language embedding for each prompt. In both the image-text retrieval and visual question answering, all data in the test set can be treated as belonging to unseen classes.

\subsection{Main Results}
\paragraph{Image Classification.~}
The experimental results for image classification are shown in Table~\ref{tab:classification}. 
With better prompts learned from counterfactual examples, our CPL method achieves clear advantages over CoCoOp for both seen and unseen classes across almost all datasets. Particularly on unseen classes, we gain an average relative improvement of $3.55\%$.

Meanwhile, CoCoOp shows its poor generalization ability. Specifically, we found that CoCoOp performs worse than CLIP on StandfordCars on both seen and unseen classes, and on Caltech101 and Flower102 on unseen classes, indicating that it tends to learn and leverage spurious relations and could not generalize well on unseen classes in some cases. We believe all these mentioned above can be sufficient evidence that
the main idea of CPL, learning non-spurious prompt representation can aid CLIP adapting at test time, is practical. 

\paragraph{Image-Text Retrieval.~}
Table~\ref{tab:ir} reports results on image-text retrieval on the unseen test set. CPL can beat the zero-shot CLIP consistently across the three different settings, demonstrating that CPL can also learn better prompt representation and more effectively exploit the limited amount of data on image-text retrieval. Meanwhile, CoCoOp performs even worse than CLIP on Flickr30k using $0.5\%$ training data, which suggests that a tiny quantity of training data for image-text retrieval can lead to spurious prompt representation if using naïve instance-conditional prompt tuning method.

\paragraph{Visual Question Answering.~}
For visual question answering, the results are shown in Table~\ref{tab:vqa}. As can be seen, CPL surpasses the baseline CoCoOp with a relative improvement of up to $25.08\%$ when using $1\%$ instances for training. This proves the concept that CPL can be effective on more complicated vision-and-language tasks. In fact, visual question answering is more challenging for zero-shot CLIP which is pre-trained for image-text matching. During pre-training, CLIP sees most sentences similar to captions in image-text retrieval and those captions can be directly used as prompts; while for VQA, question-answer pairs have to be adapted into declarative prompts. Therefore, zero-shot CLIP has poor performance on VQA, but few-shot prompt tuning via CPL can help reduce the prompt domain gap significantly. Apart from the vanilla CPL method, we examined another variant of CPL where we do not add additional category information into the prompt (denoted as CPL w/o. Category Information), the results indicate that constructing task-relevant prompts by adding categorical information contributes to the improvement. 

\begin{figure}[t]
\centering
\includegraphics[width=\linewidth]{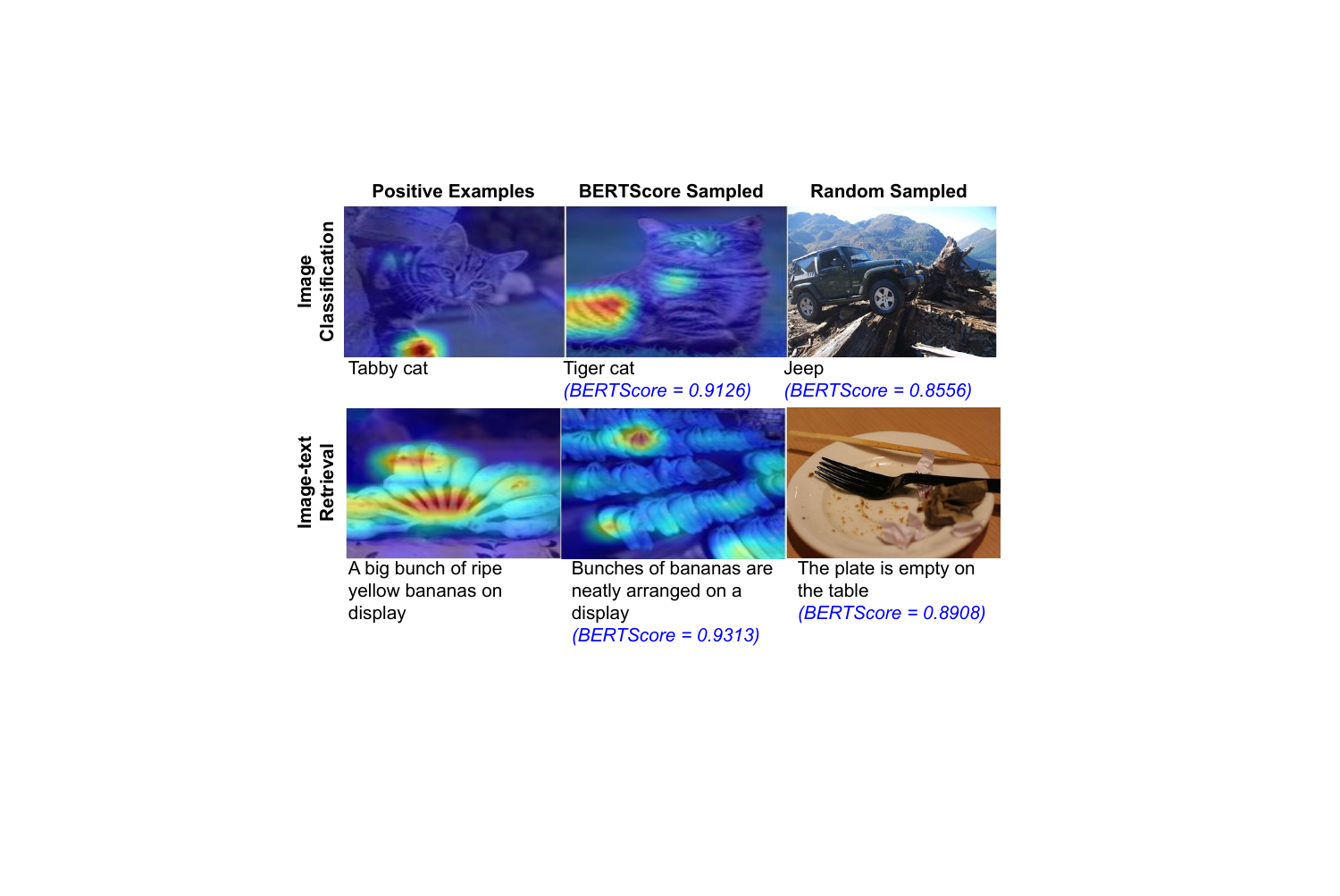}
\caption{Visualization of the weights of the controller parameter $\mathbf{u}$ on images. The first column is the original positive examples; the second column is BERT-sampled negative examples; the third column is randomly-sampled negative examples for comparison. The BERTScore between the text prompts of positive examples and sampled examples are shown at the bottom.
}
\label{fig:visualize}
\end{figure}

\subsection{Ablation Analysis}
\paragraph{Negative Sampling.~} We compare the random sampling vs. BERTScore sampling over ImageNet for image classification, MSCOCO for image-text retrieval, and VQAv2 for visual question answering in Table~\ref{tab:sample}. With more challenging negative examples, BERTScore sampling leads to more effective prompt tuning and overbeats random sampling on all three tasks. The qualitative visualizations of the two sampling strategies are shown in Figure~\ref{fig:visualize}, from which it can be seen that BERTScore-sampled images are much more semantically similar to the original images.

\paragraph{Non-spurious Feature Visualization.}
We visualize the heatmap of the learned non-spurious feature weights in the image level in Figure~\ref{fig:visualize}. The weights are mainly centralized on the semantically meaningful regions that are aligned to the text prompts. 
    
\begin{table}[t]
 \resizebox{\columnwidth}{!}{
  \centering
  \begin{tabular}{llll}
    \toprule
    Method & {ImageNet}& {MSCOCO}& VQAv2 \\
   \midrule
    Random sampling & 75.28&57.78& 33.01
\\
    BERTScore sampling &\textbf{76.02}&\textbf{58.43}&\textbf{35.66}\\
   \bottomrule
  \end{tabular}
  }
  \caption{Random sampling vs. BERTScore sampling for CPL over three tasks.  On ImageNet, we measure the average accuracy across seen and unseen classes. On MSCOCO and VQAv2, we both use 1\% instances for few-shot learning.}
 \label{tab:sample}
\end{table}

\begin{figure}[t]
\centering
\includegraphics[width=0.65\linewidth]{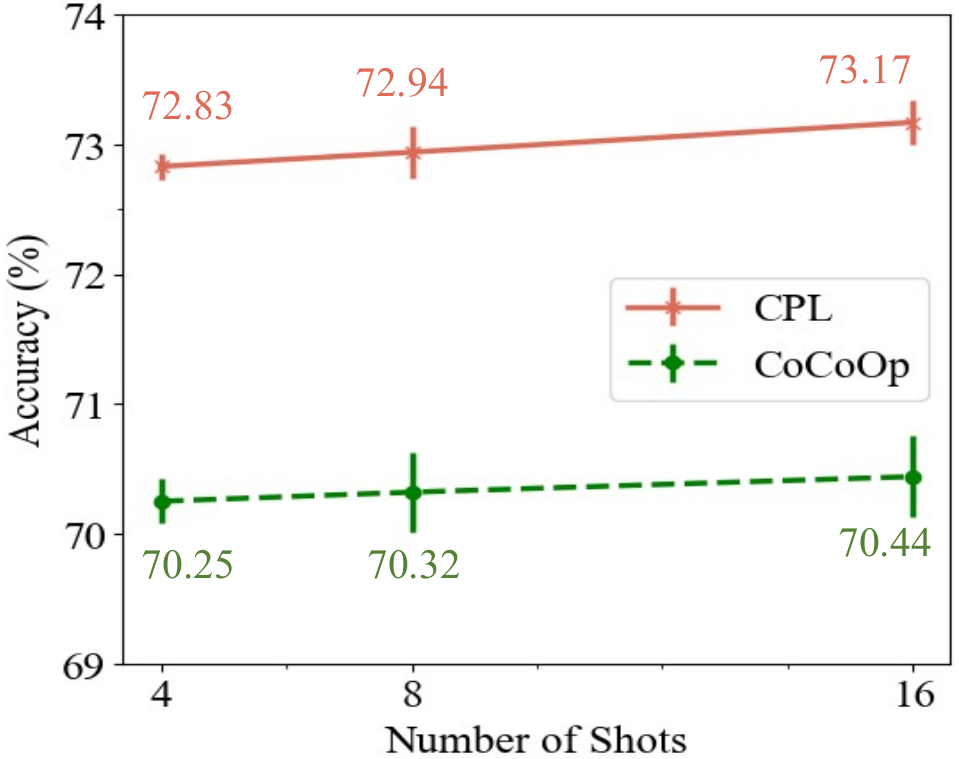}
\caption{Accuracy comparison on ImageNet~\cite{imagenet} unseen classes under three different shots. CPL performs better than CoCoOp consistently and has lower standard errors.
}
\label{fig:shots}
\end{figure}

\paragraph{Number of Shots in Image Classification.~} 
We then study the effects of the number of shots on CPL for image classification. Following the few-shot evaluation protocol adopted in CLIP, we use $4$, $8$, and $16$ shots for training on ImageNet. From Figure~\ref{fig:shots}, increasing the number of shots keeps improving the performance of both two methods on unseen classes. Meanwhile, CPL outperforms CoCoOp under the three different settings and has lower standard errors.

\begin{figure}[t]
\centering
\includegraphics[width=0.65\linewidth]{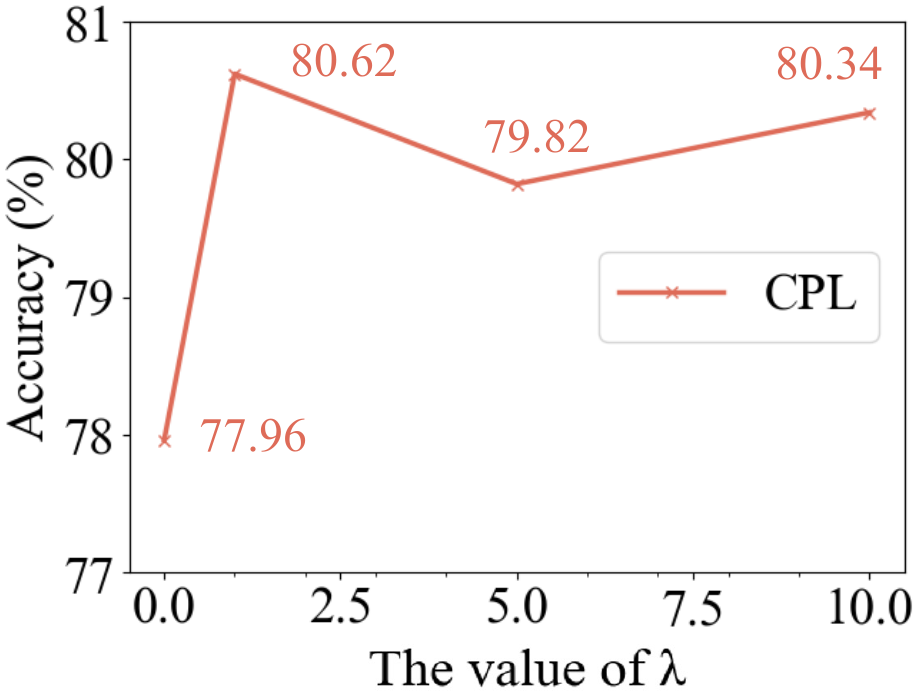}
\caption{Ablation of four different $\lambda$ values on the SUN397 dataset in terms of average accuracy (\%). The performance of CPL peaks at $\lambda =1$.}
\label{fig:lambda}
\end{figure}

\paragraph{Contribution of Contrastive Learning.} 
In Section~\ref{sec:method}, we use the coefficient $\lambda$ to weigh the contrastive learning loss and combine it with the cross-entropy loss. It is observed that the scale of contrastive learning loss is smaller, hence we try to use a larger $\lambda$ to balance the two loss terms. Figure~\ref{fig:lambda}
shows the average accuracy result across seen and unseen classes on the SUN397 dataset under four different $\lambda$ values. Note that when $\lambda$ is zero, there is no contribution from the contrastive loss and the method actually learns the prompt using standard cross-entropy loss. From experimental results obtained on the SUN397 dataset, we can observe that using $\lambda = 1$
leads to the best performance.

\section{Conclusion}
In this paper, we propose a Counterfactual Prompt Learning (CPL) framework to avoid time-consuming prompt engineering and learn more generalizable prompt representation for vision and language models.  We conduct abundant experiments on
seven widely used image classification datasets, two image-text retrieval datasets, and one visual question answering dataset. Our proposed CPL method outperforms the previous prompt tuning baseline and the zero-shot CLIP across the three tasks. In the future, we plan to develop more sophisticated methods based on CPL and extend
CPL to other vision and language tasks.

\section*{Limitations}
There are fairness issues in large pre-trained vision and language models such as CLIP. The proposed prompt learning method in this study automatically learns the prompt and does not address those issues in the pre-trained model. Considering the method is proposed for the few-shot setting, careful inspection and tuning are also needed when testing our method on other biased datasets. The methodologies proposed in~\citet{multimodal_fairness} and~\citet{clip_fairness} may possibly be paired with CPL to potentially address the issues. Another limitation is the absence of explainability in CPL, which is a common problem with existing soft prompt tuning methods. Back-mapping tuned soft prompts representation to natural language is a way for interpretation; however, due to the limited size of vocabulary used by CLIP during the training, prior methods such as searching for the nearest words in the embedding space can not accurately match the vector to natural language. Expanding the dictionary size for CLIP embedding or developing more advanced back-mapping techniques can possibly address the limitation. 

\section*{Acknowledgments}
We would like to thank the support of the Google Ads Faculty Research Award. 
We also thank the anonymous reviewers for their thought-provoking comments.
The views and conclusions contained in this document are those of the authors and should not be interpreted as representing the sponsor.

\bibliography{anthology,custom}
\bibliographystyle{acl_natbib}

\appendix
\appendix

\section{Visualization of Sampled Images}
\label{sec:visualization}
\begin{figure*}[t]
\centering
\includegraphics[width=0.8\linewidth]{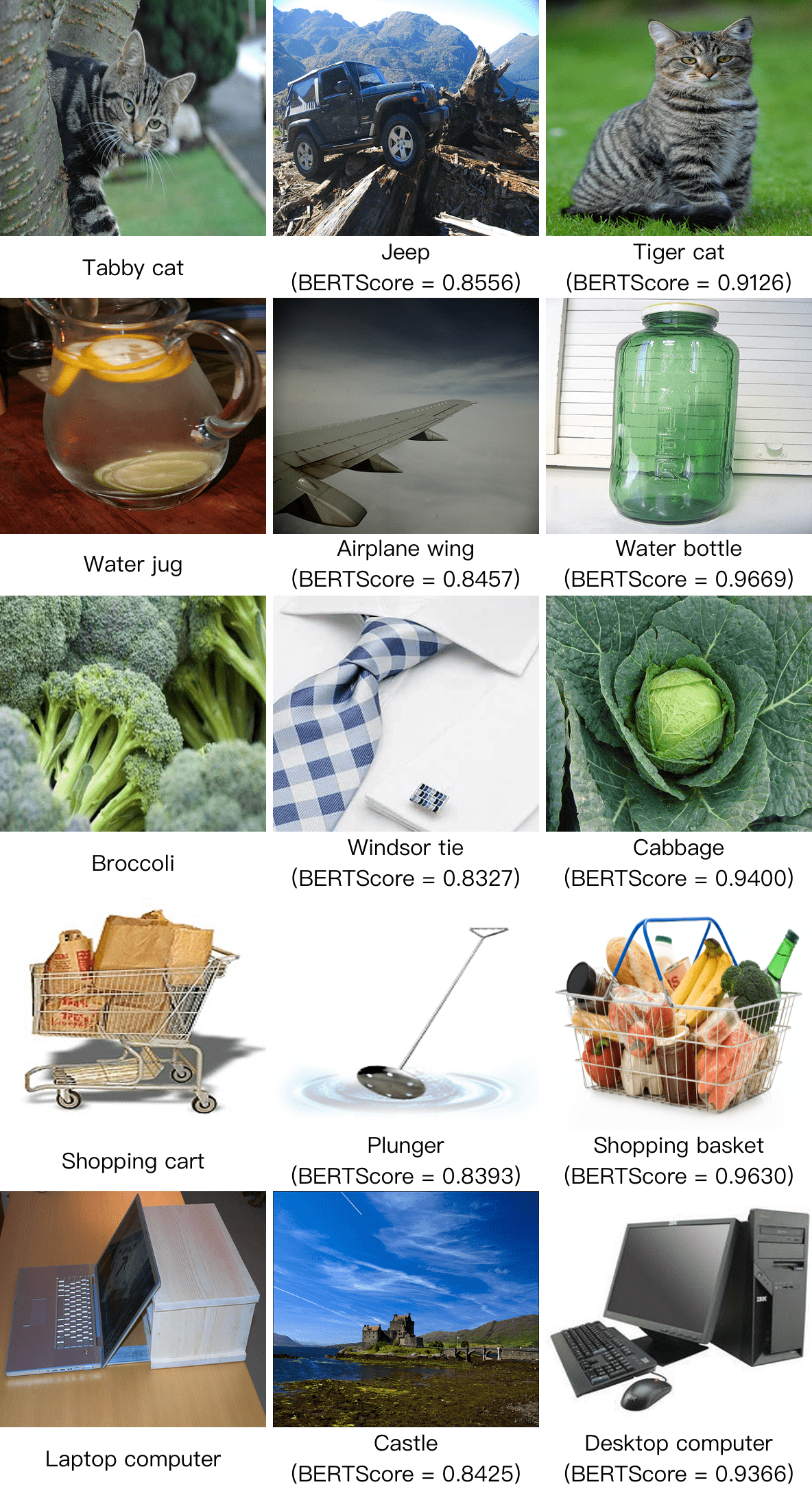}
\caption{Comparison of sampled images from the ImageNet dataset via random sampling and BERTScore sampling. The first column is original positive examples. The second column is randomly sampled images. The third column is BERTScore sampled images.}
\label{fig:icsampling}
\end{figure*}

\begin{figure*}[t]
\centering
\includegraphics[width=0.8\linewidth]{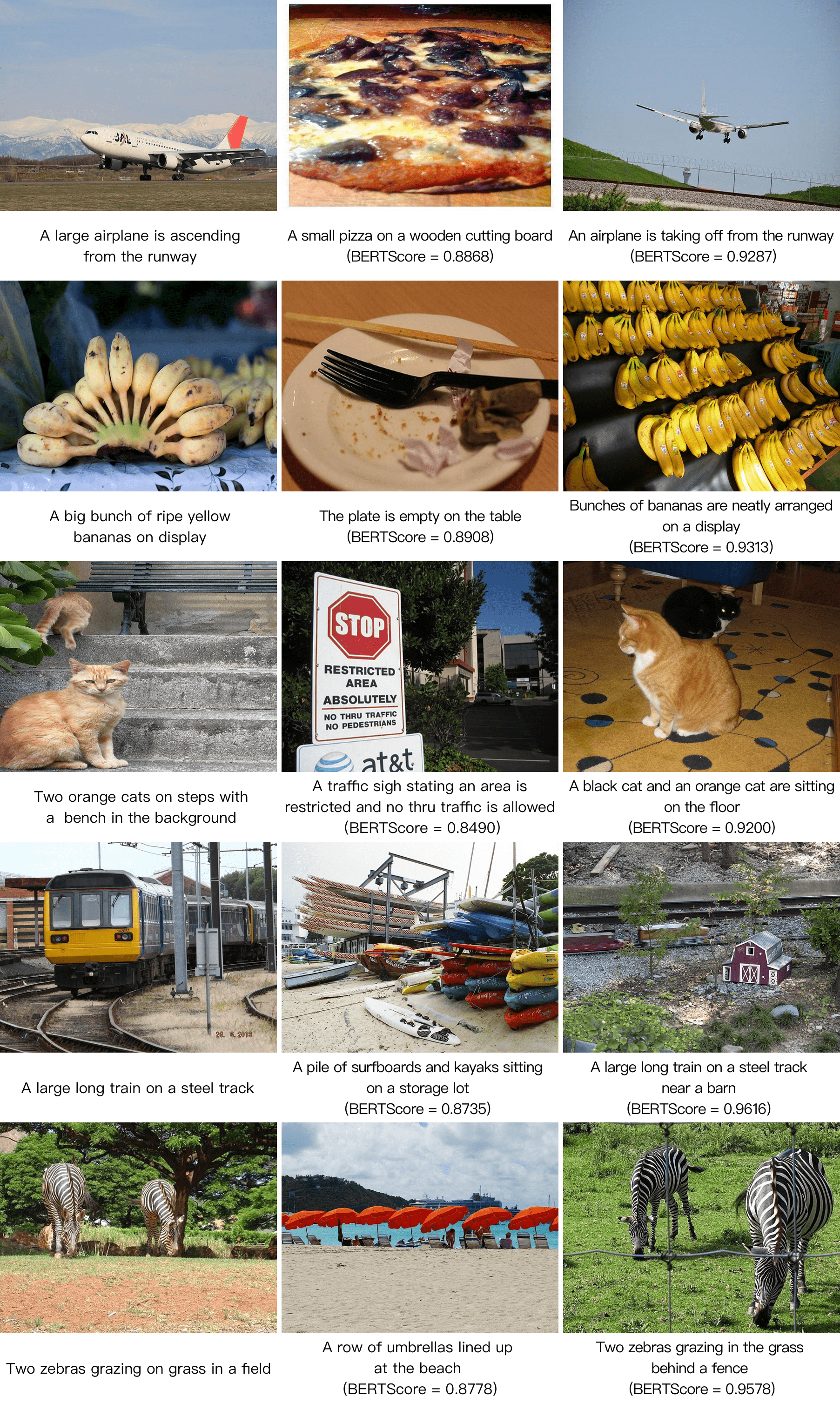}
\caption{Comparison of sampled images from the COCO dataset via random sampling and BERTScore sampling. The first column is original positive examples. The second column is randomly sampled images. The third column is BERTScore sampled images.}
\label{fig:irsampling}
\end{figure*}

\begin{figure*}[t]
\centering
\includegraphics[width=0.8\linewidth]{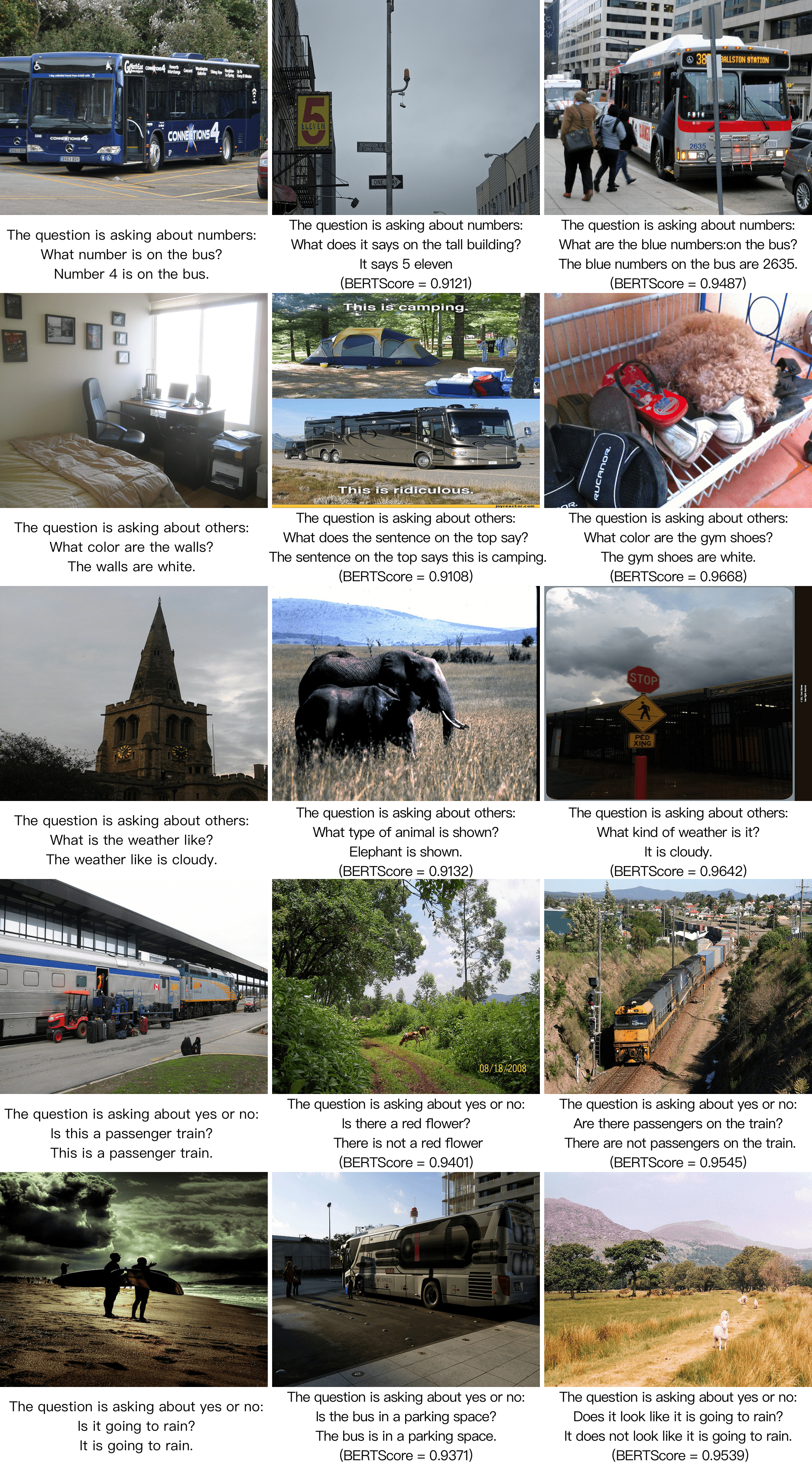}
\caption{Comparison of sampled images from the VQAv2 dataset via random sampling and BERTScore sampling. The first column is original positive examples. The second column is randomly sampled images. The third column is BERTScore sampled images.}
\label{fig:vqasampling}
\end{figure*}

We visualize the comparison of CLIP retrieved image pairs via random sampling and BERTScore sampling for image classification as shown in Figure~\ref{fig:icsampling}, image-text retrieval as shown in Figure~\ref{fig:irsampling}, and visual question answering as shown in Figure~\ref{fig:vqasampling}.

\section{Hyperparameter Settings}
Here we report the hyperparameter for different tasks in Table~\ref{tab:setting1}, Table~\ref{tab:setting2}, and Table~\ref{tab:setting3}. Most of the hyperparameters are fixed, while a small number of settings have slight changes due to the nature of the different tasks.

\section{Number of Parameters}
The model used in our experiments is CLIP using pre-trained Vision Transformer (ViT-B-224/16) and transformer initialized from BERT~\cite{bert}. The number of parameters for the model is 151M.

\begin{table}[H]
 \resizebox{0.8\columnwidth}{!}{
  \centering
  \begin{tabular}{c c}
    \toprule
    Hyperparameters & Value \\
    \midrule
    Optimizer & SGD\\
    Learning rate & 0.002\\
    LR scheduler & Cosine annealing\\
    Warmup LR & Constant 1e-5\\
    Warmup epoch & 1\\
    Warmup type & Constant\\
    CLIP vision encoder & ViT-B/16\\
    Random seed & 1\\
    \bottomrule
  \end{tabular}}
  \caption{Hyperparameters setting for Image Classification, Image-text Retrieval, and VQA using CoCoOp. }
 \label{tab:setting1}
\end{table}

\begin{table}[H]
 \resizebox{0.8\columnwidth}{!}{
  \centering
  \begin{tabular}{c c}
    \toprule
    Hyperparameters & Value  \\
    \midrule
    Optimizer & SGD\\
    Learning rate & 0.002\\
    LR scheduler & Cosine annealing\\
    Warmup LR & Constant 1e-5\\
    Warmup epoch & 1\\
    Warmup type & Constant\\
    CLIP vision encoder & ViT-B/16\\
    Random seed & 1\\
    \bottomrule
  \end{tabular}}
  \caption{Hyperparameters setting for Image Classification, Image-text Retrieval, and VQA using CPL.}
 \label{tab:setting2}
\end{table}

\begin{table}[H]
 \resizebox{0.8\columnwidth}{!}{
  \centering
  \begin{tabular}{c c}
    \toprule
    Hyperparameters & Value  \\
    \midrule
    Optimizer & SGD\\
    Learning rate & 0.002\\
    LR scheduler & Cosine annealing\\
    Warmup LR & Constant 1e-5\\
    Warmup epoch & 1\\
    Warmup type & Constant\\
    CLIP vision encoder & ViT-B/16\\
    Random seed & 1\\
    \bottomrule
  \end{tabular}}
  \caption{Hyperparameters setting for text manipulation using CoCoOp.}
 \label{tab:setting3}
\end{table}

\end{document}